\documentclass{article} % For LaTeX2e
\usepackage{iclr2024_conference,times}
\iclrfinalcopy

\usepackage{amsmath,amsfonts,bm}

\def\eqref#1{equation~\ref{#1}}
\def\1{\bm{1}}

\DeclareMathAlphabet{\mathsfit}{\encodingdefault}{\sfdefault}{m}{sl}
\SetMathAlphabet{\mathsfit}{bold}{\encodingdefault}{\sfdefault}{bx}{n}

\usepackage[colorlinks=true, allcolors=blue]{hyperref}
\usepackage{xurl}
\usepackage{cmap}
\usepackage{cleveref}
\usepackage{enumitem}
\usepackage{booktabs}
\usepackage{multirow}
\usepackage[utf8]{inputenc}
\usepackage[T1]{fontenc}
\usepackage{pifont}
\usepackage[framemethod=TikZ]{mdframed}
\usepackage{graphicx}
\usepackage{subcaption}
\newmdenv[%
    backgroundcolor=yellow!10,
    linecolor=black,
    outerlinewidth=0.5pt,
    roundcorner=1mm,
    skipabove=\topsep,
    skipbelow=\topsep,
    font=\ttfamily\small,
]{promptbox}

\title{Airavata: Introducing Hindi Instruction-tuned LLM}

\author{%\small
    \textbf{Jay Gala}$^{1}$ \quad
    \textbf{Thanmay Jayakumar}$^{1}$ \quad
    \textbf{Jaavid Aktar Husain}$^{1,3}$ \quad
    \textbf{Aswanth Kumar}$^{4}$ \vspace{0.5em} \\
    \textbf{Mohammed Safi Ur Rahman Khan}$^{1}$ \quad
    \textbf{Diptesh Kanojia}$^{5}$ \quad
    \textbf{Ratish Puduppully}$^{6}$ \vspace{0.5em} \\
    \textbf{Mitesh M. Khapra}$^{1,2}$ \quad
    \textbf{Raj Dabre}$^{7}$ \quad
    \textbf{Rudra Murthy}$^{8}$ \quad
    \textbf{Anoop Kunchukuttan}$^{1,2,9}$ \vspace{1em} \\
    $^{1}$Nilekani Centre at AI4Bharat \quad
    $^{2}$IIT Madras \quad
    $^{3}$IIIT D\&M Kancheepuram \vspace{0.5em} \\
    $^{4}$Flipkart \quad 
    $^{5}$University of Surrey \quad 
    $^{6}$A$^*$STAR \quad 
    $^{7}$NICT \quad 
    $^{8}$IBM Research \quad
    $^{9}$Microsoft
}

\newcommand{\fturl}[1]{\footnote{\small \url{#1}}}

\begin{document}

\maketitle

% \begin{abstract}
% The abstract paragraph should be indented 1/2~inch (3~picas) on both left and
% right-hand margins. Use 10~point type, with a vertical spacing of 11~points.
% The word \textsc{Abstract} must be centered, in small caps, and in point size 12. Two
% line spaces precede the abstract. The abstract must be limited to one
% paragraph.
% \end{abstract}

\begin{table}[h]
\centering
\begin{tabular}{c}
\url{https://ai4bharat.github.io/airavata}
\end{tabular}
\end{table}

\begin{figure}[ht]
    \centering
    \includegraphics[width=0.5\textwidth]{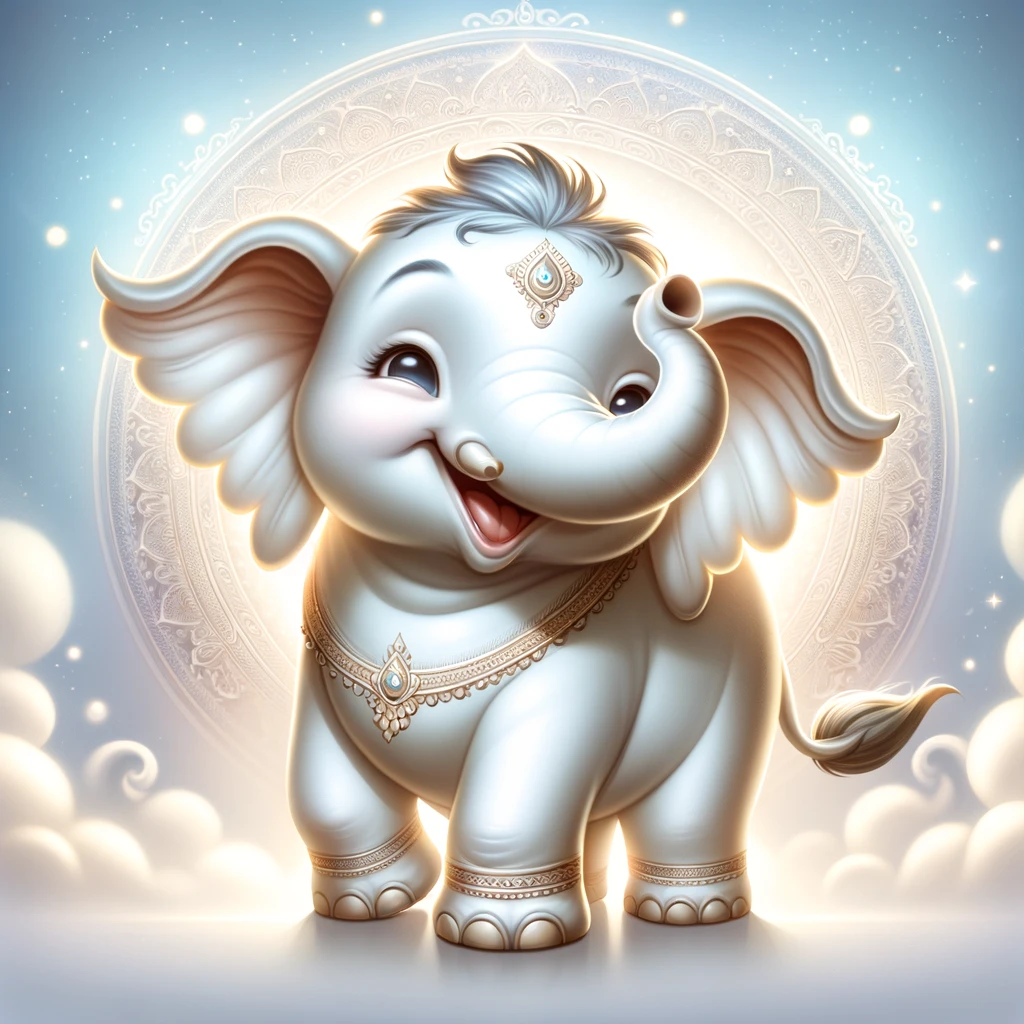}
    \caption{Image Courtesy: DALL-E 3 \citep{dalle3}.}
    \label{fig:cover}
\end{figure}

\section{Introduction}

The recent year has witnessed tremendous interest and activity in the world of Large Language Models (LLMs). LLMs hold the potential to unlock exciting applications in artificial intelligence due to their ability to comprehend complex natural language instructions and excel in a broad spectrum of tasks involving language, knowledge, reasoning, and creative generation. To foster research, innovation, and widespread adoption, an open ecosystem is essential. We have observed significant advancements in this area with the launch of models like Llama 2 \citep{touvron2023llama} and Mistral \citep{jiang2023mistral}, as well as their instruction-tuned variants such as Llama 2 Chat \citep{touvron2023llama}, Mistral-Instruct \citep{jiang2023mistral}, and Zephyr \citep{tunstall2023zephyr}, among others. Major progress has also been made in developing datasets for pre-training such as RedPajama \citep{together2023redpajama}), instruction tuning (e.g., Alpaca \citep{alpaca}, UltraChat \citep{ding2023enhancing}, Dolly \citep{DatabricksBlog2023DollyV2}, OpenAssistant \citep{köpf2023openassistant}, LMSYS-Chat \citep{zheng2023lmsyschat1m}), and evaluation benchmarks (e.g., AlpacaEval \citep{dubois2023alpacafarm}, MT-Bench \citep{zheng2023judging}). However, most of these advancements have been predominantly centered around the English language.

There is limited support for Indian languages, which can be attributed to the incidental inclusion of some Indian language data that slipped through the data filters during the pre-training of these language models. However, the representation of data, the efficacy of tokenizers, and task performance for Indian languages are considerably behind that of English. The performance in Indian languages, even on closed-source models such as ChatGPT \citep{chatgpt}, GPT-4 \citep{openai2023gpt4}, and others, is inferior compared to English \citep{ahuja-etal-2023-mega}. Therefore, there is an urgent need to develop a similar ecosystem of tools, models, and datasets for Indian languages to foster research and innovation. In pursuit of this objective, the recent collaboration with Sarvam AI led to release of OpenHathi \citep{sarvam2023openhathi}, an open-source foundational model for Hindi, developed by extending Llama 2 \citep{touvron2023llama}.

Today, we announce the next step -- an initial release of ``Airavata'', an instruction-tuned model for Hindi built upon finetuning OpenHathi \citep{sarvam2023openhathi} with diverse, instruction-tuning Hindi datasets to make it better suited for assistive tasks.

Along with the model, we also share the instruction tuning datasets\fturl{https://huggingface.co/datasets/ai4bharat/indic-instruct-data-v0.1} to enable further research for IndicLLMs. We rely on human-curated, license-friendly instruction-tuned datasets to build ``Airavata''. We do not use data generated from proprietary models like GPT-4 \citep{openai2023gpt4}, etc. We think this is a more sustainable way of building instruction-tuned models at scale for most Indic languages, where relying on distilled data from commercial models would increase costs and restrict their free usage in downstream applications due to licensing restrictions.

We also compile a collection of evaluation benchmarks\footnote{\url{https://huggingface.co/collections/ai4bharat/airavata-evaluation-suite-65b13b7b68165de71ba0b333}} along with an evaluation framework to compare various LLMs for their abilities on diverse tasks when instructed in Hindi. Using this benchmark and human judgments, we compare different LLMs to quantify the current state of their Hindi capabilities. We conduct a detailed analysis of Airavata's performance on various Natural Language Understanding (NLU) and Natural Language Generation (NLG) tasks and find that the instruction fine-tuning helps align the model to various NLU tasks. There is significant potential for improvement in NLG tasks, which require the creation of larger, more diverse instruction datasets as well as innovations in aligning English model representations to Hindi representations to drive better cross-lingual transfer.

\section{Instruction Tuning Dataset Creation}

High-quality instruction tuning datasets are important for the good performance of LLMs. However, there are few diverse datasets for Hindi. Following \cite{wei2023polylm}, we rely on translating high-quality English-supervised instruction-tuning datasets into Hindi. We use IndicTrans2 \citep{gala2023indictrans}, the state-of-the-art open-source MT model for Indian languages, for translation. Some previous works \citep{li2023bactrianx,wei2023polylm} have used ChatGPT \citep{chatgpt} to translate instructions and/or generate responses into Hindi to better use context during translation (IndicTrans2 and most MT models are sentence-level). However, this is not cost-effective, and the translation quality of ChatGPT \citep{chatgpt} is lower than IndicTrans2 \citep{gala2023indictrans}, and its generation quality in Hindi might not be up to the mark \citep{ahuja-etal-2023-mega}. A future avenue of work would be improving translation quality when document context is available.

We sample examples from different datasets listed in \Cref{tab:instruction-datasets} to ensure balanced representations across all the tasks while fitting into our instruction tuning budget. We translate the instructions, input, and outputs into Hindi. This results in a total of 404k examples spanning English and Hindi language. The translated Hindi examples were filtered to retain high-quality examples. Specifically, examples were retained only when the chrF++ score \citep{popovic-2017-chrf} between the back-translated example and the corresponding English example was 50 or above. The final dataset used for instruction tuning contains 385k examples. \Cref{tab:instruction-datasets} shows the details of the final training dataset. The dataset is available on the HuggingFace Hub.\fturl{https://huggingface.co/datasets/ai4bharat/indic-instruct-data-v0.1}

\begin{table}[]
\centering
\begin{tabular}{lrrrr}
\toprule
\multirow{2}{*}{Dataset} & \multicolumn{2}{c}{Unfiltered} & \multicolumn{2}{c}{Filtered} \\
\cmidrule{2-5}
                         & English        & Hindi         & English       & Hindi        \\
\midrule
FLAN-v2 \citep{longpre2023flan}  & 67,463         & 67,463        & 67,463        & 65,228       \\
Anthropic-HHH \citep{bai2022training}  & 5,000          & 5,000         & 5,000         & 4,911        \\
Dolly \citep{DatabricksBlog2023DollyV2}  & 15,011         & 15,011        & 15,011        & 14,880       \\
OpenAssistant  \citep{köpf2023openassistant} & 19,945         & 20,128        & 19,945        & 16384        \\
LymSys-Chat \citep{zheng2023lmsyschat1m}  & 50,000         & 50,000        & 50,000        & 37,422       \\
WikiHow                  & 20,400         & 6,055         & 20,400        & 6,055        \\
Anudesh                  & 5,234          & 7,577         & 5,234         & 7,577       \\
\midrule
Total       &   183,053     &   171,234	    &   183,053     &	152,457 \\
\bottomrule
\end{tabular}
\caption{Instruction Fine-tuning Training Dataset Details}
\label{tab:instruction-datasets}
\end{table}

We also create two native Hindi Instruction datasets:

\begin{itemize}[leftmargin=*]
\item \textbf{wikiHow:} wikiHow\fturl{https://www.wikihow.com/Main-Page} is an online wiki-style platform that serves as a valuable resource for a diverse array of how-to articles spanning numerous topics. The articles on the platform are human-moderated, ensuring a high-quality standard. The questions users pose in these articles closely align with potential use cases for this model, making it a rich resource for training models. Additionally, this might also help induce reasoning capabilities and generate logical step-by-step responses. We curate around 20k and 6k articles in English and Hindi, respectively, resulting in a total of around 27k articles. We currently formulate the data as a completion task given either question or question along with a few initial steps. The dataset is released under the CC-0 license.
\item \textbf{Anudesh:} Anudesh is a crowd-sourced collection of prompts accompanied by responses generated from the Llama 2 70B model \citep{touvron2023llama}. Participants are provided with clear guidelines detailing the nature of the interaction required, including the specific language to be employed. These languages encompass a range that includes Indic languages, English, transliterated Indic, as well as a blend of Indic and English in a code-mixed format. Contributors craft their prompts in adherence to these directives and the specified language criteria. Subsequently, these prompts are then paired with the corresponding translated outputs from the Llama 2 70B model \citep{touvron2023llama}. The dataset is released under CC-BY-4.0 license. More details about the interactions will be released soon.
\end{itemize}

We provide a brief description of all the existing instruction datasets used in addition to the above (and corresponding licenses) below:

\begin{itemize}[leftmargin=*]
\item \textbf{FLAN-v2} \citep{longpre2023flan}: A collection of NLP tasks that combines a number of existing NLP datasets with various data augmentations, introduced by \cite{chung2022scaling}. We sample around 67K examples for our training mixture. The dataset is released under the Apache-2.0 license.
\item \textbf{Anthropic-HHH} \citep{bai2022training}: A collection of human-collected preference data for aligning the models to be helpful and harmless. We sample 5K conversations from the ``chosen" column for our training mixture. The dataset is released under the MIT license.
\item \textbf{Dolly} \citep{DatabricksBlog2023DollyV2}:	A corpus of more than 15K records generated by thousands of Databricks employees to enable LLMs to exhibit the magical interactivity of ChatGPT. The dataset is released under the CC-BY-SA-3.0 license.
\item \textbf{OpenAssistant} \citep{köpf2023openassistant}: A collection of human-generated, human-annotated assistant-style conversation corpus consisting of 38K messages, resulting in over 3K conversation trees and around 20K conversations. The dataset is released under the Apache-2.0.
\item \textbf{LymSys-Chat} \citep{zheng2023lmsyschat1m}: A collection of 1M real-world conversations spanning 25 SOTA LLMs similar to OpenAssistant \citep{köpf2023openassistant}. We sample 50K conversations for our training mixture. The dataset is released under the LMSYS-Chat-1M Dataset License Agreement.\fturl{https://huggingface.co/datasets/lmsys/lmsys-chat-1m\#lmsys-chat-1m-dataset-license-agreement}
\item \textbf{NMT} \citep{gala2023indictrans}: A multi-domain human-annotated dataset containing 50K bitext English-Hindi translation pairs from BPCC-Human \citep{gala2023indictrans} to enable better cross-lingual transfer. The dataset is released under the CC-BY-4.0 license.
\end{itemize}

\section{Supervised Fine-tuning}

We fine-tune the OpenHathi model using the above-compiled datasets. We perform parameter-efficient fine-tuning with LoRA \citep{hu2022lora}. The hyperparameters used are listed in the \Cref{tab:hyperparameters}.

\begin{table}[t]
\centering
\begin{tabular}{ll}
\toprule
Hyper-Parameter          & Value                                                                         \\
\midrule
LoRA Rank                & 16                                                                            \\
LoRA alpha               & 32                                                                            \\
LoRA Dropout             & 0.05                                                                          \\
LoRA Target Modules      & q\_proj, v\_proj, k\_proj, gate\_proj \\
&  up\_proj, down\_proj \\
Epochs                   & 4                                                                             \\
Learning rate            & 5e-4                                                                          \\
Batch Size               & 128                                                                           \\
Floating Point Precision & bfloat16      \\
\bottomrule
\end{tabular}
\caption{Hyperparameters for Fine-tuning}
\label{tab:hyperparameters}
\end{table}

During fine-tuning, the loss was computed only for the output tokens. We used the OpenInstruct framework\fturl{https://github.com/allenai/open-instruct} for fine-tuning and customizing it for our requirements (our custom version is available as IndicInstruct\fturl{https://github.com/AI4Bharat/IndicInstruct}). One fine-tuning example corresponds to one example in the dataset. However, this is suboptimal since many tokens are wasted as padding tokens. We plan to optimize this process by packing multiple dataset examples into a single fine-tuning example \citep{iyer2022optiml,krell2023efficient}.

\subsection{Full vs. LoRA finetuning}

\begin{figure}[t]
    \centering
    \includegraphics[width=\textwidth]{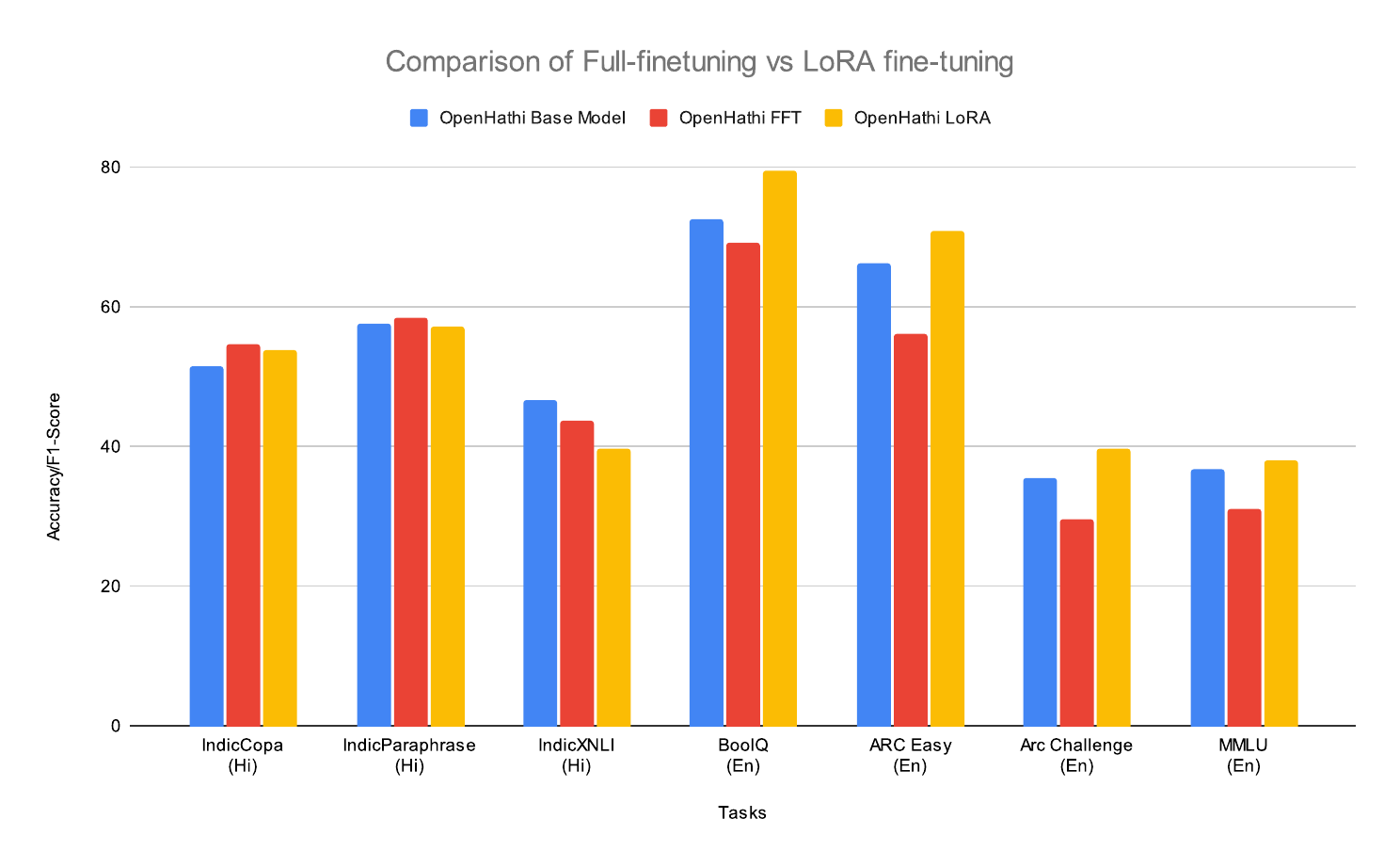}
    \caption{Ablation experiment to understand the performance gaps between Full fine-tuning and LoRA fine-tuning across a mix of English and Hindi NLU tasks.}
    \label{fig:fft_vs_lora_abl}
\end{figure}

There are two prevalent methods for instruction fine-tuning for LLMs: Full fine-tuning and LoRA. Full fine-tuning (FFT) involves updating all the model parameters, whereas LoRA \citep{hu2022lora} fine-tuning involves introducing a small set of additional parameters and only updating them. 
We perform an ablation study to examine both the fine-tuning strategies to train two models, one employing full fine-tuning and the other by LoRA fine-tuning. For the context of this ablation study, we consider the FLAN v2 \citep{longpre2023flan} subset from our instruction dataset spanning English and Hindi. For our evaluation, we used a subset of NLU tasks in Hindi, along with BoolQ \citep{clark2019boolq} and MMLU \citep{hendrycks2021measuring} tasks in English, as development sets to decide between full fine-tuning and LoRA fine-tuning (\Cref{fig:fft_vs_lora_abl}). We observed that FFT models outperform the OpenHathi base model \citep{sarvam2023openhathi} in IndicCopa \citep{doddapaneni2022towards} and IndicXParaphrase \citep{doddapaneni2022towards} tasks. However, the FFT model performed poorly on English tasks compared to the base model as well as the LoRA fine-tuned model. LoRA fine-tuned model demonstrated improvements or similar performance as the OpenHathi base model on both Hindi NLU and English tasks. As a result, we decided to proceed with LoRA fine-tuning for training our final model. We report results on the LoRA fine-tuned final model in the subsequent section.

\subsection{Model Selection}

We fine-tune the OpenHathi model for 4 epochs and save the model after each epoch. We evaluate the checkpoint of each epoch on the dev set (IndicSentiment, IndicCOPA, IndicXNLI, and IndicQA from IndicXTREME \citep{doddapaneni2022towards}, and Flores devtest \citep{flores101,nllb2022}) and compare the average performance. We observe that the checkpoint of epoch 3 performs well on NLU tasks, while the checkpoint of epoch 4 performs well on NLG tasks. We perform checkpoint averaging, where we interpolate the weights of the above two checkpoints to obtain a model that performs well across both NLU and NLG tasks. We found the best interpolation factor to be around 0.6.

\begin{equation*}
    \mathrm{interpolated \hspace{2pt} weights} = 0.6 \times \mathrm{checkpoint_3} + (1 - 0.6) \times \mathrm{checkpoint_4}
\end{equation*}

\section{Evaluation on NLP Benchmarks}

We evaluate our model on the standard NLU and NLG benchmarks, which include native Hindi test sets from IndicXTREME \citep{doddapaneni2022towards} and Indic NLG Suite \citep{kumar-etal-2022-indicnlg}. Further, to test the knowledge and reasoning capabilities of the model in Hindi, we translate the English benchmarks such as MMLU \citep{hendrycks2021measuring}, Hellaswag \citep{zellers-etal-2019-hellaswag}, ARC \citep{clark2018think}, Winogrande \citep{sakaguchi2020winogrande} and BoolQ \citep{clark2019boolq} and use these to evaluate our final model. The IndicTrans2 \citep{gala2023indictrans} model was employed for translating the benchmarks. Although this is not the perfect evaluation setup however, due to limited resources, we rely on this method as a proxy indicator for assessing trends in the performance of the model in terms of reasoning capabilities in Hindi. An important future direction would involve creating equivalent benchmarks in the native language instead of solely relying on translations.

\subsection{Results}

\Cref{tab:indic-nlu-results,tab:english-nlu-results,tab:translation-results,tab:indic-nlg-results} shows the comparison of Airavata with the base model (OpenHathi) and the translate-test baseline using the Llama 2 7B Chat model. In the translate-test approach, the Hindi input is translated into English using the IndicTrans2 model and is used as a prompt to the Llama 2 7B Chat model. We observe that Airavata significantly outperforms the OpenHathi model on most tasks, demonstrating that fine-tuning on the IndicInstruct dataset helps align the base model to a variety of tasks. Moreover, the performance of the translate-test approach with the Llama 2 7B Chat model has a lot of variance, while Airavata consistently performs well across tasks. OpenHathi and Airavata exhibit comparable performance in the translation task. OpenHathi benefits from its extensive training on parallel corpora, resulting in a highly proficient base model for the translation task. We observe mixed results on generation tasks, indicating the scope for further improvement of Airvata, especially in open-ended text generation capabilities. \Cref{tab:english-nlu-results} shows the performance of Airavata and other models on English test sets and the respective machine-translated Hindi test set. We observe a 5-15 point performance gap between the English baseline and the respective Hindi baseline across various tasks for both OpenHathi and Airavata models. This indicates limited cross-lingual transfer of English knowledge to Hindi, highlighting the need for future research to improve cross-lingual alignment to maximize knowledge transfer between English and Hindi.

\begingroup
\setlength{\tabcolsep}{3pt} % Default value: 6pt
\renewcommand{\arraystretch}{1} % Default value: 1
\begin{table}[h]
\small
\centering
\resizebox{\textwidth}{!}{%
\begin{tabular}{lcccccc}
\toprule
 & \multicolumn{3}{c}{\textbf{0-Shot}} & \multicolumn{3}{c}{\textbf{5-Shot}} \\
\cmidrule{2-4} \cmidrule{5-7}
 & \multirow{2}{*}{\textbf{OpenHathi}} & \textbf{Llama2 7B Chat} & \multirow{2}{*}{\textbf{Airavata}} & \multirow{2}{*}{\textbf{OpenHathi}} & \textbf{Llama2 7B Chat} & \multirow{2}{*}{\textbf{Airavata}} \\
 &  & \textbf{(translate-test)} &  &  & \textbf{(translate-test)} &  \\
\midrule
IndicSentiment & 72.89 & \textbf{97.85} & 95.81 & 96.59 & \textbf{98.43} & 97.01 \\
IndicCopa & 68.69 & \textbf{76.53} & 63.75 & 42.77 & \textbf{78.34} & 72.97 \\
IndicXNLI & 16.67 & 23.67 & \textbf{73.26} & 42.25 & 47.96 & \textbf{74.70} \\
IndicXParaphrase & 71.72 & 09.54 & \textbf{76.53} & 66.67 & 48.56 & \textbf{69.87} \\
\bottomrule
\end{tabular}%
}
\caption{F1 scores on Indic NLU and Commonsense Reasoning tasks}
\label{tab:indic-nlu-results}
\end{table}
\endgroup

\begingroup
\setlength{\tabcolsep}{3pt} % Default value: 6pt
\renewcommand{\arraystretch}{1} % Default value: 1
\begin{table}[h]
\small
\centering
\begin{tabular}{llcccc}
\toprule
\multicolumn{1}{c}{} & \multicolumn{1}{c}{\multirow{2}{*}{\textbf{Variant}}} & \multicolumn{2}{c}{\textbf{0-Shot}} & \multicolumn{2}{c}{\textbf{5-Shot}} \\
\cmidrule{3-4} \cmidrule{5-6}
\multicolumn{1}{c}{} & \multicolumn{1}{c}{} & \textbf{OpenHathi} & \textbf{Airavata} & \textbf{OpenHathi} & \textbf{Airavata} \\
\midrule
\multirow{2}{*}{MMLU} & English & 36.16 & \textbf{41.39} & 40.12 & \textbf{43.28} \\
 & Hindi (Translated) & 32.27 & \textbf{34.96} & 35.13 & \textbf{36.00} \\
 \midrule
\multirow{2}{*}{BoolQ} & English & 52.63 & \textbf{73.00} & \textbf{64.46} & 62.02 \\
 & Hindi (Translated) & 58.56 & \textbf{64.50} & \textbf{65.69} & 51.47 \\
 \midrule
\multirow{2}{*}{ARC Easy} & English & 57.28 & \textbf{70.50} & 62.12 & \textbf{71.04} \\
 & Hindi (Translated) & 44.28 & \textbf{54.00} & 49.87 & \textbf{54.84} \\
 \midrule
\multirow{2}{*}{Arc Challenge} & English & 39.85 & \textbf{45.90} & 46.25 & \textbf{48.29} \\
 & Hindi (Translated) & 32.68 & \textbf{35.92} & \textbf{36.60} & 36.26 \\
 \midrule
\multirow{2}{*}{Hella Swag} & English & 31.48 & \textbf{34.37} & 32.45 & \textbf{36.83} \\
 & Hindi (Translated) & \textbf{25.59} & 25.37 & \textbf{24.85} & \textbf{24.84} \\
 \midrule
Winogrande & English & 49.17 & \textbf{49.72} & \textbf{-} & \textbf{-} \\
\bottomrule
\end{tabular}
\caption{Accuracy on English NLU and Commonsense Reasoning tasks and its translated variants}
\label{tab:english-nlu-results}
\end{table}
\endgroup

\begingroup
\setlength{\tabcolsep}{3pt} % Default value: 6pt
\renewcommand{\arraystretch}{1} % Default value: 1
\begin{table}[!ht]
\small
\centering
\begin{tabular}{lccccc}
\toprule
\multicolumn{1}{c}{\multirow{2}{*}{}} & \multirow{2}{*}{\textbf{Metric}} & \multicolumn{2}{c}{\textbf{0-Shot}} & \multicolumn{2}{c}{\textbf{5-shot}} \\
\cmidrule{3-4} \cmidrule{5-6}
\multicolumn{1}{c}{} &  & \textbf{OpenHathi} & \textbf{Airavata} & \textbf{OpenHathi} & \textbf{Airavata} \\
\midrule
\multirow{2}{*}{Flores} & chrF++ & \textbf{55.41} & 54.82 & \textbf{54.98} & 54.24 \\
\cmidrule{2-6}
 & BLEURT & \textbf{0.7103} & 0.6970 & \textbf{0.7115} & \textbf{0.7084} \\
\midrule
\multirow{2}{*}{IN22-Gen} & chrF++ & \textbf{54.23} & 53.78 & \textbf{54.53} & 52.81 \\
\cmidrule{2-6}
& BLEURT & \textbf{0.7156} & 0.7012 & \textbf{0.7181} & 0.7037 \\
\bottomrule
\end{tabular}
\caption{chrF++ and BLEURT scores on English-Hindi translation task}
\label{tab:translation-results}
\end{table}
\endgroup

\begingroup
\setlength{\tabcolsep}{3pt} % Default value: 6pt
\renewcommand{\arraystretch}{1} % Default value: 1
\begin{table}[!ht]
\small
\centering
\begin{tabular}{lcccc}
\toprule
\multirow{3}{*}{} & \multirow{3}{*}{\textbf{Metric}} & \multicolumn{3}{c}{\textbf{1-shot}} \\
\cmidrule{3-5}
 &  & \multirow{2}{*}{\textbf{OpenHathi}} & \textbf{Llama 2 7B Chat} & \multirow{2}{*}{\textbf{Airavata}} \\
 &  &  & \textbf{(translate-test)} &  \\
\midrule
Indic QA (No Context) & F1 & 17 & 4.58 & \textbf{21.01} \\
\midrule
Indic QA (With Context) & F1 & 20.69 & 19.59 & \textbf{37.69} \\
\midrule
\multirow{2}{*}{Indic Headline} & Rouge L & 11.26 & \textbf{23.06} & 12.32 \\
\cmidrule{2-5}
 & BLEURT & \textbf{0.4682} & 0.4590 & 0.3793 \\
 \midrule
\multirow{2}{*}{IndicWikiBio} & Rouge L & 20.45 & \textbf{41.01} & 10.66 \\
\cmidrule{2-5}
 & BLEURT & 0.5185 & \textbf{0.6415} & 0.4279 \\
\bottomrule
\end{tabular}
\caption{F1, Rouge L and BLEURT scores on Indic NLG tasks}
\label{tab:indic-nlg-results}
\end{table}
\endgroup

\section{Human Evaluation}

We assess Airavata using authentic prompts provided by real users, evaluating its performance across five distinct abilities outlined in \Cref{tab:abilities-desc}.

\begingroup
\setlength{\tabcolsep}{3pt} % Default value: 6pt
\renewcommand{\arraystretch}{1} % Default value: 1
\begin{table}[h]
\small
\centering
\begin{tabular}{ll}
\toprule
\textbf{AbilityName} & \textbf{Ability} \\
\midrule
\textbf{Long} & Ability to generate long-form text like writing essays, speeches, reports, etc. \\
\midrule
\textbf{Fact-Ops} & Ability to give factual opinions and explanations like seeking recommendations, \\
& seeking advice, opinions, explanations, etc. \\
\midrule
\textbf{Content} & Ability to make content accessible like summarizations, layman \\
& explanations, etc \\
\midrule
\textbf{Lang-Creativity} & Ability to be creative in language like finding anagrams, rhyming words,  \\
& vocabulary enhancement, etc \\
\midrule
\textbf{Culture} & Ability to answer questions related to Indian Culture. \\
\bottomrule
\end{tabular}
\caption{Description of abilities to test through human evaluation}
\label{tab:abilities-desc}
\end{table}
\endgroup

We define a set of intents and domains of focus, which are then provided to users, along with clear instructions on the type of prompts they should construct. More details about the benchmark will be released soon.

Along with Airavata, we also evaluate ChatGPT \citep{chatgpt}, GPT-4 \citep{openai2023gpt4} and BactrianX-llama-7B \citep{li2023bactrianx} models for the same abilities. BactrianX-llama-7B is an instructed fine-tuned model for Hindi directly fine-tuned with the Llama base model on a multilingual instruction dataset. This multilingual instruction dataset consisted of machine-translated instructions from Alpaca \citep{alpaca} and Dolly \citep{DatabricksBlog2023DollyV2} datasets, followed by response generation from ChatGPT \citep{chatgpt}. Annotators were presented with a prompt and a randomly selected response from any of the models and were instructed to provide ratings based on the metrics outlined in Table \ref{tab:evaluation-rubrics}.

\begingroup
\setlength{\tabcolsep}{3pt} % Default value: 6pt
\renewcommand{\arraystretch}{1} % Default value: 1
\begin{table}[]
\small
\centering
\begin{tabular}{p{0.2\linewidth}p{0.6\linewidth}c}
\toprule
\multicolumn{1}{c}{\textbf{Metric}} & \multicolumn{1}{c}{\textbf{Details}} & \textbf{Range} \\
\midrule
\textbf{IFA: Instruction Following Ability} & This assesses the model's ability to accurately and effectively follow the instructions provided in the prompt & 0-2 \\
\midrule
\textbf{CNS: Closeness to Native Speaker} & This assesses how naturally and fluently the model’s responses align with the way a native Hindi speaker would express the same ideas. & 0-2 \\
\midrule
\textbf{CQ: Content Quality} & This evaluates the response in terms of its factual accuracy, logical flow of ideas, and overall informational relevance. & 0-2 \\
\bottomrule
\end{tabular}
\caption{Rubrics for Human Evaluation}
\label{tab:evaluation-rubrics}
\end{table}
\endgroup

In addition to the above metrics, we also ask the human evaluators to give a final score on the likert scale of 1 to 5, reflecting their overall satisfaction with the response.

\begin{figure}[t]
    \centering
    \includegraphics[width=\textwidth]{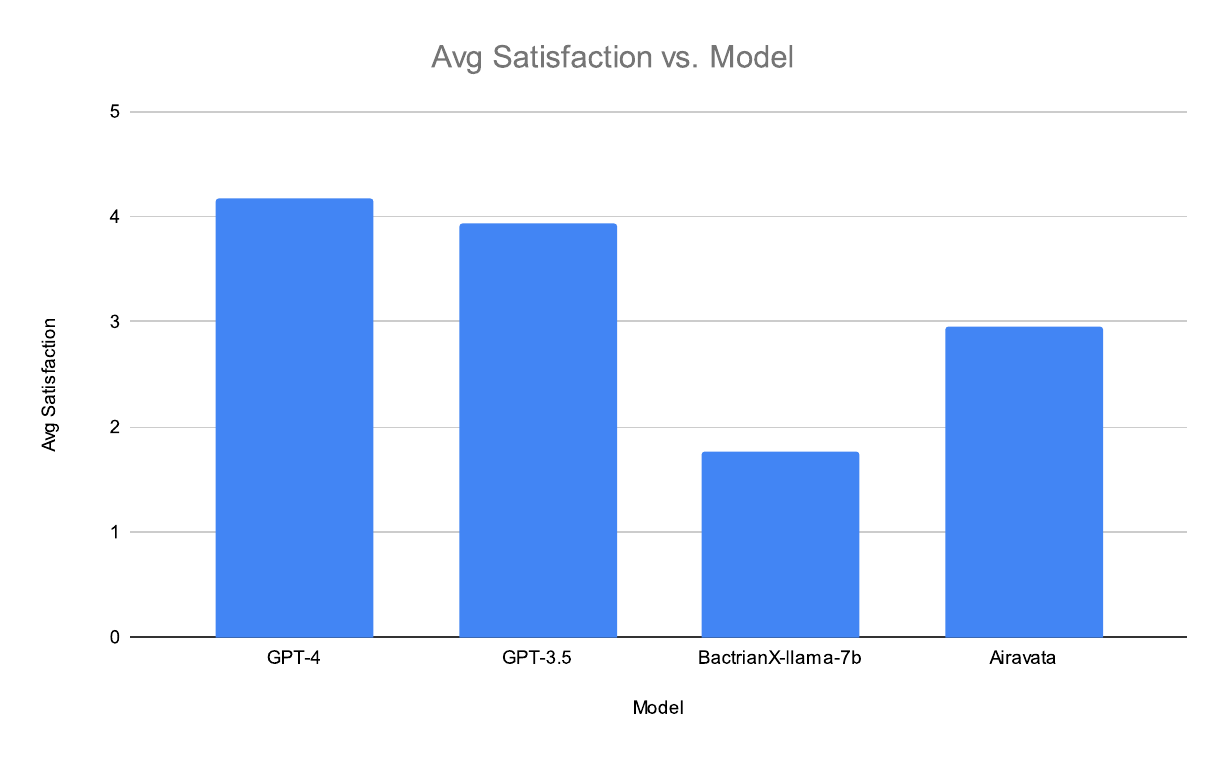}
    \caption{Average satisfaction scores for various models based on a Likert scale between 1 and 5 reported by Human annotators.}
    \label{fig:avg_satisfaction}
\end{figure}

\begin{figure}[t]
    \centering
    \includegraphics[width=\textwidth]{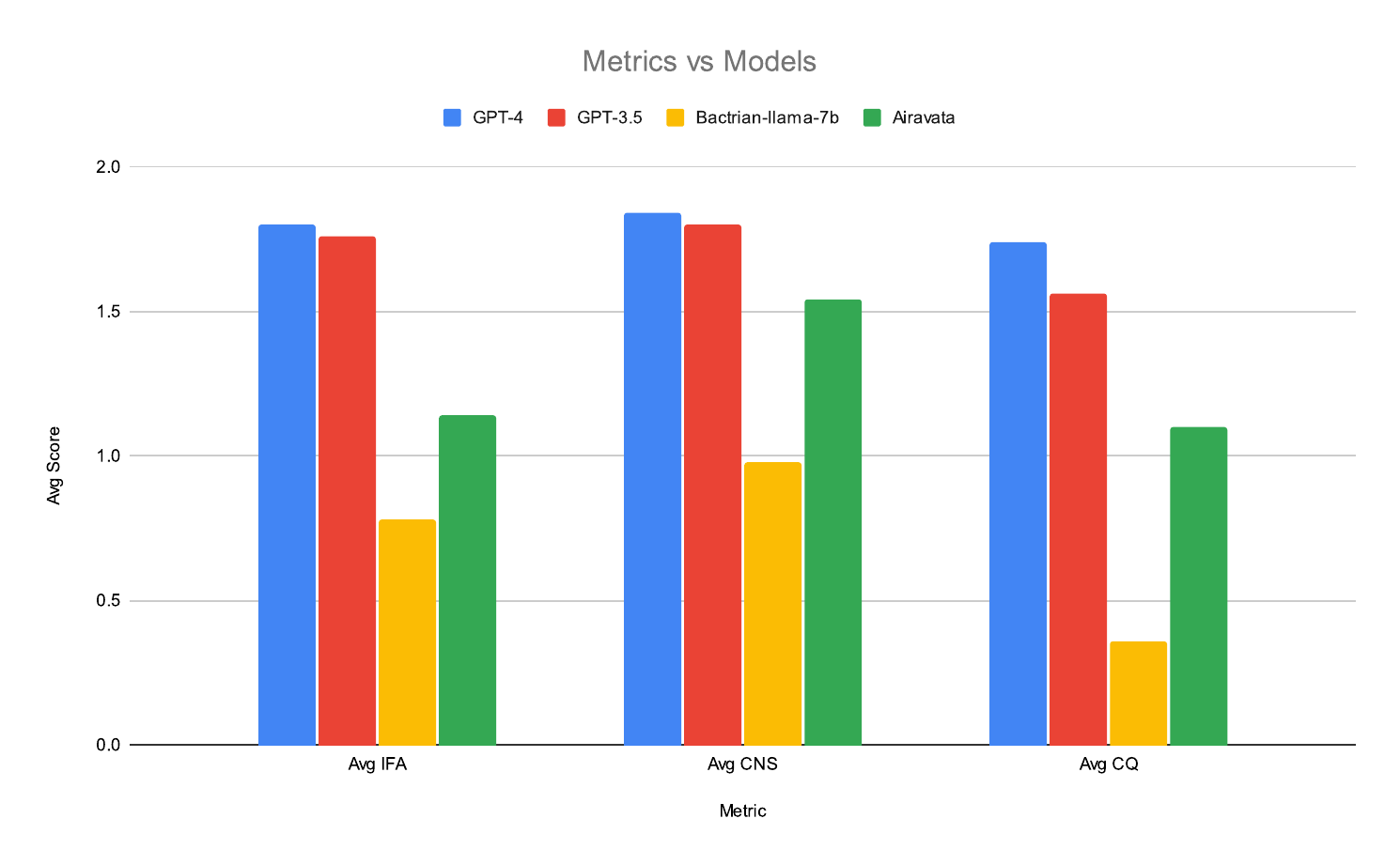}
    \caption{Human evaluation scores for assessing the instruction following and content generation abilities of the models based on the rubrics described in \Cref{tab:evaluation-rubrics}.}
    \label{fig:metrics_vs_model}
\end{figure}

\begin{figure}[t]
    \centering
    \includegraphics[width=\textwidth]{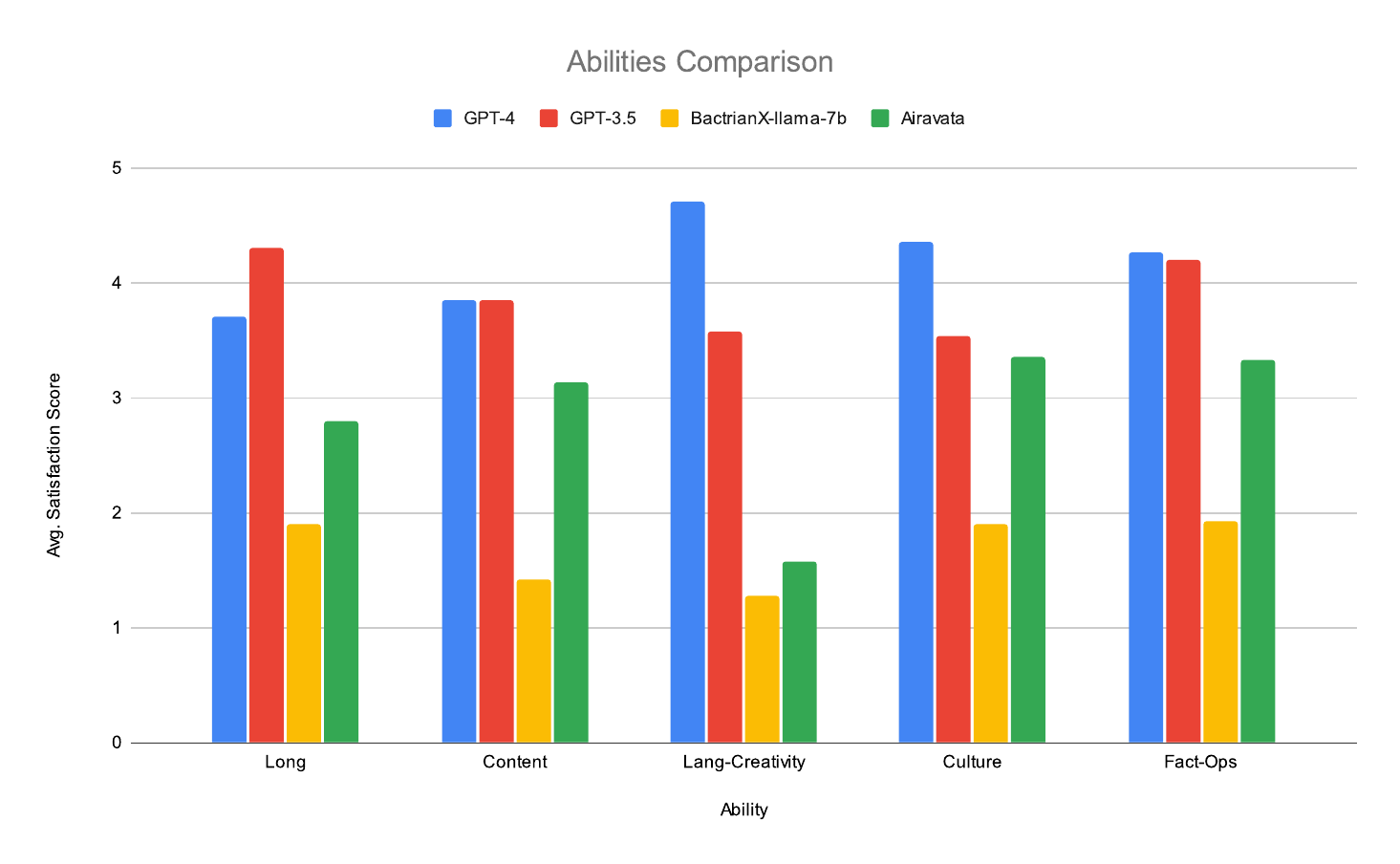}
    \caption{Fine-grained human evaluation of content generation abilities of the models described in \Cref{tab:abilities-desc}.}
    \label{fig:abilities_comparison}
\end{figure}

We sample a set of 50 prompts\fturl{https://huggingface.co/datasets/ai4bharat/human-eval} covering various intents and domains (more details about the benchmark will be made available soon) and generate the responses from all three models. The prompt-response pairs were shuffled randomly and anonymized, ensuring no indication of the generating model, before being presented to the annotators for evaluation. Annotators were instructed to assess responses strictly adhering to the aforementioned rubrics. We report the various results in \Cref{fig:avg_satisfaction,fig:metrics_vs_model,fig:abilities_comparison}.

Our observations suggest that while Airavata still trails significantly behind GPT-4 in terms of its ability to follow instructions and the quality of its content, it performs relatively better when generating natural-sounding Hindi content compared to both GPT-4 and ChatGPT. Notably, Airavata outperforms the BactrianX-llama-7B model by a significant margin. This difference in performance can be attributed to a lack of vocabulary expansion in BactrainX-llama-7B to accommodate additional Hindi tokens and its lack of continual pre-training in Hindi. Furthermore, BactrainX-llama-7B may be trained on a lower-quality dataset for instruction tuning that was completely generated using ChatGPT. The OpenHathi model and Airavata, however, have addressed these shortcomings and show substantial improvements over the BactrainX-llama-7B model. We discuss these performance trends across various capabilities in more detail below.

The findings indicate that amongst all abilities, Airavata particularly excels in providing factual opinions and explanations, as demonstrated by the earlier examples. However, the model struggles with tasks demanding creative language usage, as our SFT data lacks components emphasizing creativity. When comparing GPT-4 and ChatGPT (GPT-3.5) performance, GPT-4 generally surpasses the latter in knowledge-intensive or creativity-demanding tasks. Nevertheless, it's noteworthy that ChatGPT either outperforms or is comparable in tasks emphasizing language generation, such as long-form content creation, offering factual opinions, and ensuring content accessibility.

We acknowledge that our current human evaluation is not robust and thorough due to the limited number of prompts in our set and each prompt and response pair being evaluated by only one annotator. However, our evaluation still offers valuable preliminary insights that will inform our efforts to improve the model. Expanding the instruction dataset to include content covering a diverse range of abilities can help improve the model's capabilities. At the same time, it is important to acknowledge that a significant portion of knowledge stems from English, which possesses the most extensive knowledge repository. Therefore, achieving better cross-lingual alignment between Hindi and English representations is essential for accurately answering factual questions and minimizing erroneous responses.

\section{Toxicity and Misinformation}
We evaluate Airavata, OpenHathi, and Llama2-7B models with publicly available benchmark datasets, in both 0-shot and 5-shot settings. Our evaluation provides insights into key dimensions for LM safety. Multilingual HateCheck (MHC) is a suite of functional tests for hate speech detection and we use its Hindi subset~\citep{das-etal-2022-hatecheckhin} to evaluate and compare the performance of all models. We translate the TruthfulQA, Implicit Hate, and a human-evaluated subset of the Toxigen dataset, to Hindi. This subset of Toxigen has been denoised to retain instances that have annotation agreement from all annotators~\citep{hosseini2023empirical}. While the Implicit Hate dataset~\citep{hosseini2023empirical} helps evaluate the model performance on detecting the subtle and implicit forms of hate speech, human-evaluated Toxigen data contains instances that are directed towards various demographics. We evaluate the model performance on detection of toxicity in these three datasets, and their translated instances using the accuracy metric. Further, for evaluating the model's capability towards answering factual questions, we use the TruthfulQA dataset~\citep{lin-etal-2022-truthfulqa} which contains multiple choice questions which are factual and can mimic common human falsehoods. 

Given the accuracy scores from our evaluation, in Table~\ref{tab:toxmisinfo}, Airavata is able to detect openly expressed hate in Hindi statements from MHC with an accuracy similar to the other two models, with similar performance in both 0- and 5-shot settings. On the more challenging instances which contain implicitly veiled hate speech, Airavata is able to identify hate with significantly better accuracy than the other two models within the translated Hindi instances. On the original Implicit Hate dataset, Llama2-7B seems to perform better, given a few examples. On the Translated Toxigen subset, Llama2-7B is able to detect targeted toxic instances against certain demographics with the highest accuracy among all three models. However, given a few examples, we observe a significant performance dip for Llama2-7B and Airavata outperforms it marginally. We observe similar performance on the original English dataset and note that Airavata is better at detecting targeted hate in Hindi, as compared to implicitly veiled hate speech. Its performance at detecting targeted hate is surprisingly better than detecting openly expressed hate speech from MHC. On the TruthfulQA dataset, in both 0- and 5-shot settings, Llama2-7B outperforms OpenHathi and Airavata. On the translated TruthfulQA data, a marginal dip in the performance can be observed which indicates that we need further investigation into the model's capability for generating misinformation. 

Overall, these results may suggest that LLMs are able to identify toxicity and hateful speech, we think that further investigation is needed to evaluate toxicity and the presence of social biases within the content generated by LLMs. In the future, we plan to investigate additional existing benchmarks and novel evaluation measures to test LLMs for content safety and reliability.

\begin{table}[]
\resizebox{\textwidth}{!}{%
\begin{tabular}{@{}llcccccc@{}}
\toprule
 &  & \multicolumn{3}{c}{\textbf{0-Shot}} & \multicolumn{3}{c}{\textbf{5-Shot}} \\ \cmidrule(l){3-8} 
 & \multicolumn{1}{c}{\textbf{Variant}} & \textbf{OpenHathi} & \begin{tabular}[c]{@{}c@{}}\textbf{Llama2 7B Chat}\\ (translate-test)\end{tabular} & \textbf{Airavata} & \textbf{OpenHathi} & \begin{tabular}[c]{@{}c@{}}\textbf{Llama2 7B Chat}\\ (translate-test)\end{tabular} & \textbf{Airavata} \\ \midrule
\textbf{Multilingual HateCheck} & Hindi & 70.15 & \textbf{70.24} & \textbf{70.24} & 70.15 & \textbf{70.24} & \textbf{70.25} \\ \midrule
 \multirow{2}{*}{\textbf{Implicit Hate}} & English & 50.65 & 57.92 & \textbf{62.33} & 51.41 & \textbf{65.02} & 62.44 \\ 
& Hindi (Translated) & 52.45 & 53.21 & \textbf{61.15} & 49.99 & 52.98 & \textbf{58.84} \\ \midrule
\textbf{Toxigen} & English & 44.91 & \textbf{83.35} & 78.63 & 42.71 & 66.34 & \textbf{72.24} \\
(human evaluated) & Hindi (Translated) & 47.75 & \textbf{83.97} & 78.56 & 42.83 & 73.20 & \textbf{74.80} \\ \midrule
\textbf{TruthfulQA} & English & 30.72 & \textbf{37.25} & 33.60 & 30.72 & \textbf{37.25} & 33.64 \\
(averaged MC1 \& MC2) & Hindi (Translated) & 34.31 & \textbf{35.66} & 35.32 & 34.31 & \textbf{35.66} & 35.32 \\ \bottomrule
\end{tabular}%
}
\caption{Accuracy on hate and toxicity identification, and answering factual questions. }
\label{tab:toxmisinfo}
\end{table}

\section{Resources}

You can find all the details about the project in this section. We release the following resources to facilitate further research in instruction-tuning for Indian language LLMs.

\begin{itemize}[leftmargin=*]
    \item \href{https://github.com/AI4Bharat/IndicInstruct}{Airavata Github Repository (for fine-tuning and evaluation)}
    \item \href{https://huggingface.co/datasets/ai4bharat/indic-instruct-data-v0.1}{Airavata SFT Dataset Collection: Huggingface}
    \item \href{https://huggingface.co/collections/ai4bharat/airavata-evaluation-suite-65b13b7b68165de71ba0b333}{Airavata Evaluation Dataset Collection: Huggingface}
\end{itemize}

\section{Summary and Future Outlook}

We release Airavata, an open-source, instruction-tuned Hindi model demonstrating promising performance across various tasks compared to existing open-source models. We provide both datasets and models to facilitate further research in improving Hindi LLMs. This marks an initial stride towards developing high-quality, open-source LLMs for Indian languages that require multi-faceted efforts ranging from the creation of extensive pre-training datasets, diverse instruction tuning datasets, and high-quality foundational models.

\section{Limitations}
Airavata, like other LLMs, faces common challenges such as potential hallucinations, resulting in the generation of fabricated information, and may struggle with accuracy in complex or specialized topics. There is also a risk of generating objectionable or biased content. Its understanding of cultural nuances and effectiveness in mixed-language contexts may be constrained. Furthermore, the model's performance is closely tied to the quality and scope of its training data, influencing its efficacy and reliability. It is crucial to note that this model is designed for research purposes and is not recommended for any production use-cases.

\clearpage

\bibliography{iclr2024_conference}
\bibliographystyle{iclr2024_conference}

\clearpage

\appendix

\section{Examples}

In this section, we show example generations from Airavata. Note that the model's outputs in examples \#A5 and \#A6 did not contain newlines between bullet points. They have been added here for readability.  

\subsection{Examples where the Airavata model generates good output}

Airavata excels in specific content generation tasks, such as soliciting mail drafts (\#A1) and addressing general how-to inquiries (\#A5, \#A6). Furthermore, it demonstrates proficiency in adhering to instructions related to output structure (\#A3 and \#A5).

\begin{figure}[h]
    \centering
    \includegraphics[width=0.95\textwidth]{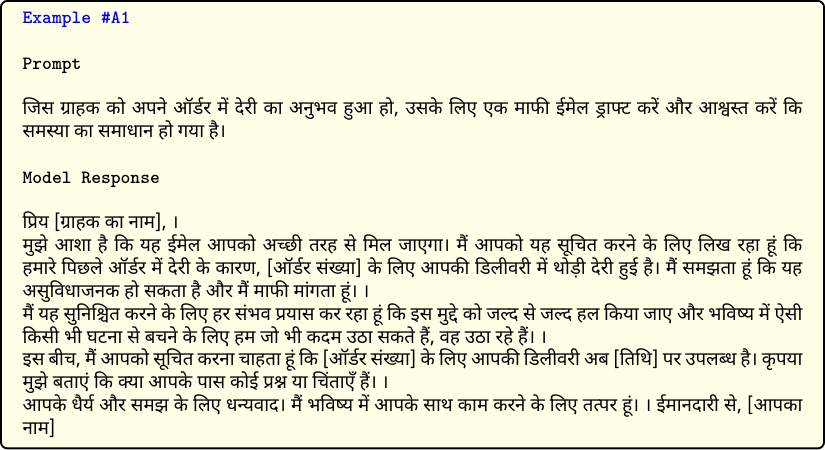}
    \label{fig:example1}
\end{figure}

\begin{figure}[h]
    \centering
    \includegraphics[width=0.95\textwidth]{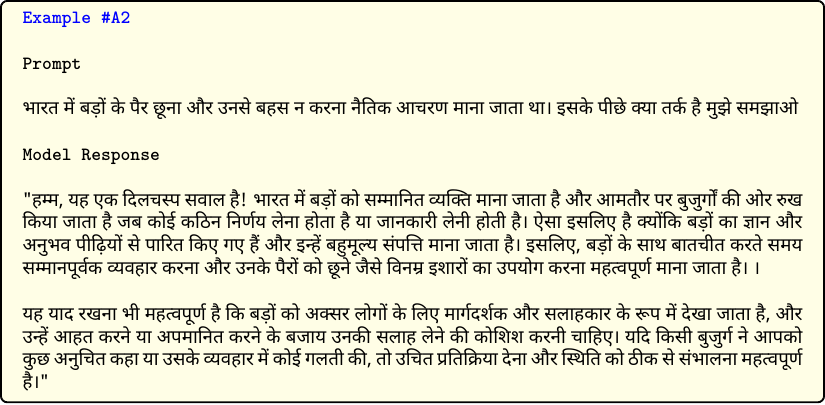}
    \label{fig:example2}
\end{figure}

\vfill

\clearpage

\begin{figure}[t]
    \centering
    \includegraphics[width=0.95\textwidth]{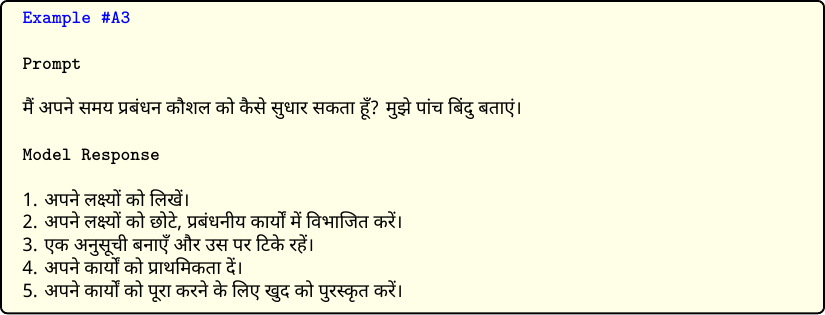}
    \label{fig:example3}
\end{figure}

\begin{figure}[h]
    \centering
    \includegraphics[width=0.95\textwidth]{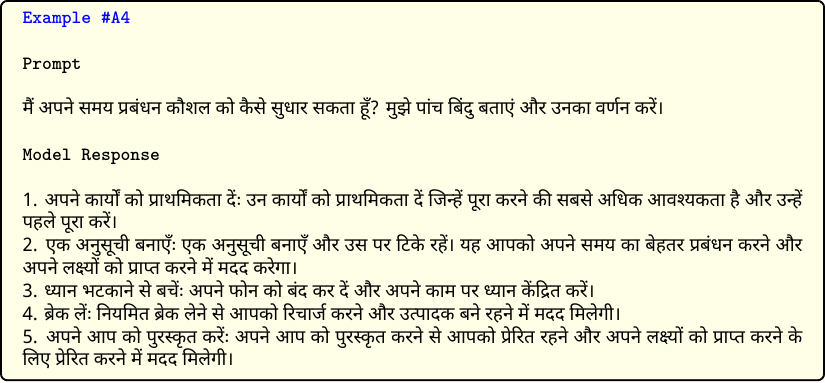}
    \label{fig:example4}
\end{figure}

\vfill

\clearpage

\begin{figure}[t]
    \centering
    \includegraphics[width=0.95\textwidth]{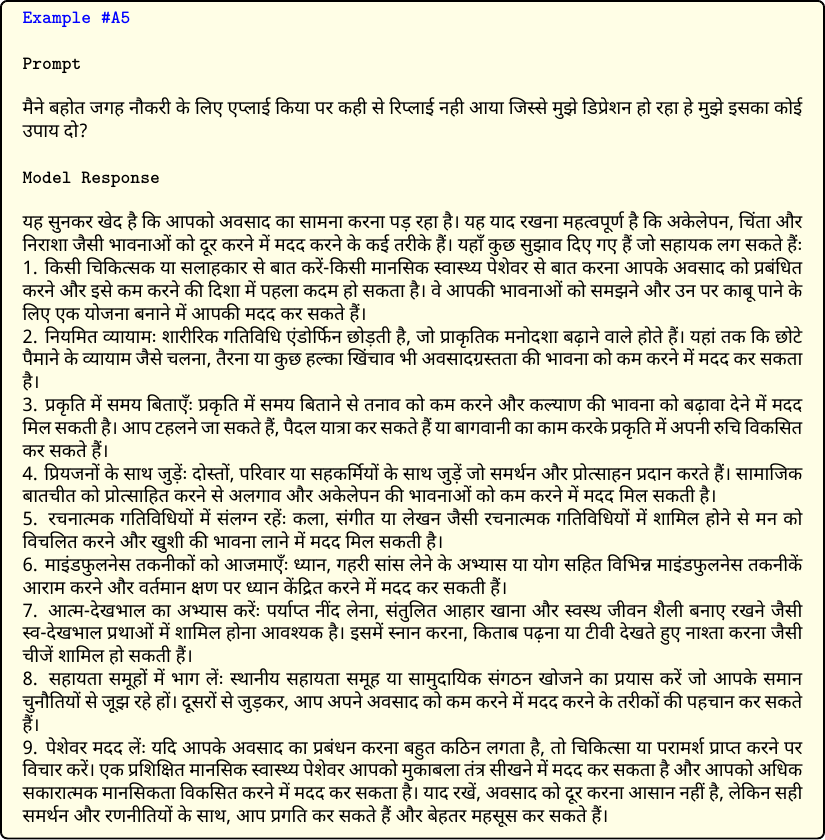}
    \label{fig:example5}
\end{figure}

\vfill

\clearpage

\begin{figure}[t]
    \centering
    \includegraphics[width=0.95\textwidth]{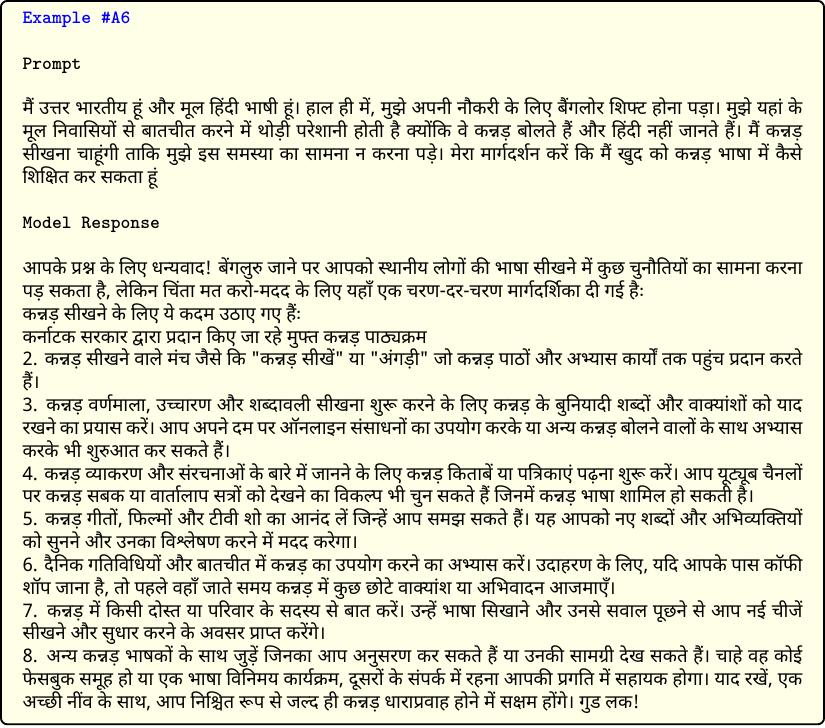}
    \label{fig:example6}
\end{figure}

\vfill

\clearpage

\subsection{Examples where Airavata output has errors}

The model is prone to hallucinations and can present distorted facts as against the plot of the novel in \#B1 or may fail to follow the instructions provided (\#B2).

\begin{figure}[h]
    \centering
    \includegraphics[width=0.95\textwidth]{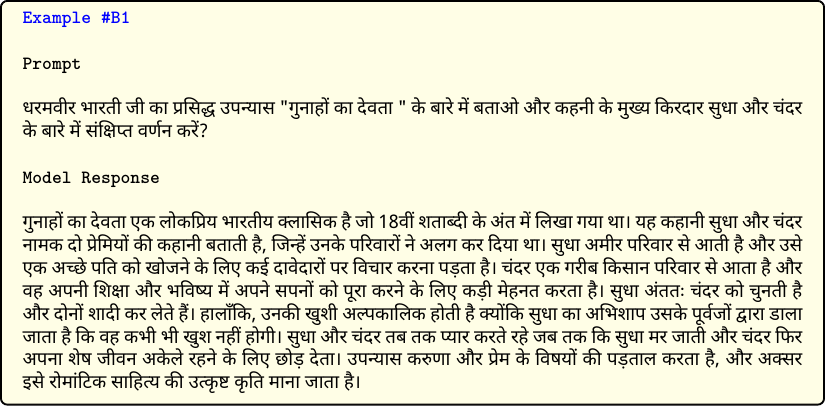}
    \label{fig:example7}
\end{figure}

\begin{figure}[h]
    \centering
    \includegraphics[width=0.95\textwidth]{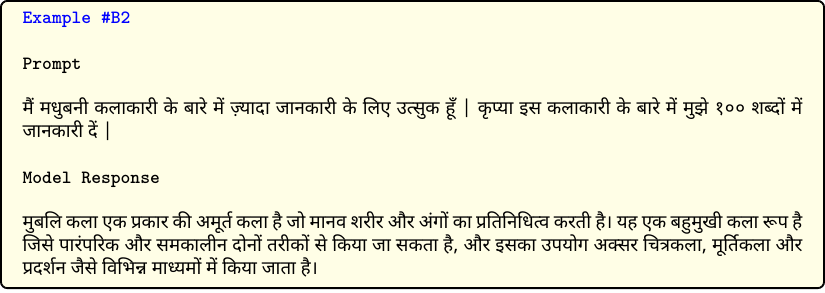}
    \label{fig:example8}
\end{figure}

\end{document}